\documentclass{article} 
\usepackage{iclr2027_conference,times}

\usepackage{amsmath,amsfonts,bm}

\def\eqref#1{equation~\ref{#1}}

\def\1{\bm{1}}

\DeclareMathAlphabet{\mathsfit}{\encodingdefault}{\sfdefault}{m}{sl}
\SetMathAlphabet{\mathsfit}{bold}{\encodingdefault}{\sfdefault}{bx}{n}

\usepackage{hyperref}
\usepackage{url}
\usepackage{graphicx}
\usepackage{booktabs}
\usepackage{makecell}
\usepackage[table]{xcolor}
\usepackage{tikz}
\usepackage{capt-of}
\usetikzlibrary{shapes.geometric} \usepackage{caption}
\definecolor{BestGreen}{RGB}{198,239,206}
\definecolor{SecondYellow}{RGB}{255,242,204}
\definecolor{GroupGray}{RGB}{242,242,242}
\newcommand{\best}[1]{\cellcolor{BestGreen}\textbf{#1}}
\newcommand{\second}[1]{\cellcolor{SecondYellow}#1}

\title{SurfSVR: 2D Surface Priors as 3D Geometric Regularizers for Sparse Voxel Reconstruction}

\author{
Yan Di$^{1}$ \quad
Chengxi Li$^{2}$ \quad
Yaoxing Wang$^{3}$ \quad
Mengge Liu$^{2}$ \quad
Zhigang Li$^{2}$ \\
\textbf{
Ruida Zhang$^{2}$ \quad
Mingyang Li$^{4}$ \quad
Pengyuan Wang$^{5}$ \quad
Shan Gao$^{3}$ \quad
Xiangyang Ji$^{2}$} \\
\small{$^1$HIT (Shenzhen) \quad $^2$Tsinghua University \quad $^3$NWPU} \\
\small{$^4$Beijing Institute of Control and Engineering \quad $^5$Technical University of Munich} \\
}

\iclrfinalcopy 
\begin{document}

\maketitle

\begin{abstract}
Sparse voxel reconstruction offers an efficient representation for high-fidelity 3D modeling, yet its geometry is commonly optimized from local photometric evidence and discrete visibility statistics. 
This often leads to fragmented surfaces, excessive subdivision, and floating artifacts, particularly in weakly textured or sparsely observed regions. 
We introduce SurfSVR, a novel sparse voxel reconstruction paradigm that treats 2D surface priors as explicit 3D geometric regularizers. 
Instead of directly lifting noisy pixel-wise depth predictions, SurfSVR first organizes each image into coherent surface regions by jointly reasoning over appearance, monocular depth, normals and cross-view geometry. 
Each region is then represented by an adaptively selected planar or quadratic surface model based on fitting reliability and geometric complexity, while cross-model agreement distinguishes reliable geometry from ambiguous predictions.
These structured 2D priors are lifted into 3D and integrated throughout the reconstruction pipeline. 
They guide surface-adaptive voxel subdivision, provide region-level depth and normal supervision during optimization, enhance geometrically reliable sparse-observed surfaces in voxel pruning, and suppress off-surface floaters during post-refinement training. 
This unified design converts semantic and geometric coherence in image space into persistent structural constraints in 3D. 
Extensive experiments on 3 public benchmarks demonstrate that SurfSVR consistently improves sparse voxel reconstruction across scenes with substantially different visibility and geometry characteristics, achieving state-of-the-art reconstruction quality. 
Codes and models will be released soon.
\end{abstract}

\begin{figure}[h]
\begin{center}
\includegraphics[width=0.9\linewidth]{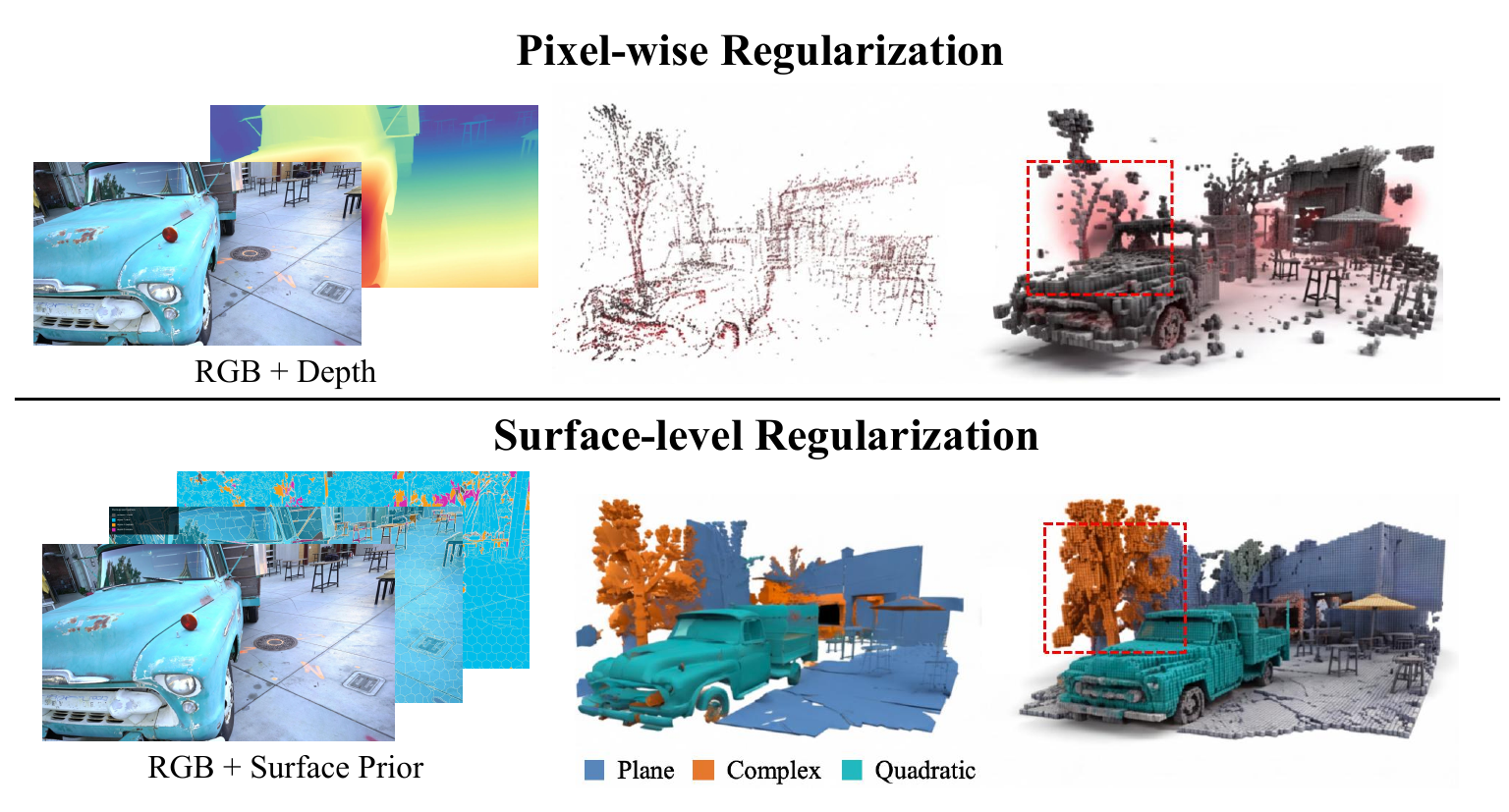}
\end{center}
\caption{
{SurfSVR achieves more complete and accurate geometry by treating 2D surface priors as explicit 3D regularizers.} Compared to state-of-the-art method~\cite{li2025geosvr}, our approach better reconstructs weakly textured surfaces, preserves thin structures, and suppresses floating artifacts.}
\label{fig:teaser}
\end{figure}

\section{Introduction}
\label{sec:intro}

High-fidelity 3D reconstruction from multi-view images is fundamental to robotics, autonomous systems, and digital content creation. 
Sparse voxel representations are particularly attractive since they combine explicit spatial organization, efficient rendering, and direct surface extraction~\cite{yu2021plenoxels,sun2022direct,li2025geosvr}. However, their geometry is still primarily optimized using local photometric evidence, rendering weights, and discrete visibility statistics. 
Such signals become unreliable in weakly textured, occluded, or sparsely observed regions, often resulting in fragmented surfaces, unnecessary subdivision, and floating artifacts. 
Moreover, visibility-based pruning introduces an inherent trade-off: conservative rules retain floaters, whereas aggressive rules remove valid thin or sparsely observed surfaces.

Recent foundation models provide informative depth, normal, and semantic cues~\cite{yang2024depthanythingv2,wang2025vggt}. 
Nevertheless, directly lifting pixel-wise predictions into 3D is unreliable since predicted geometry can be scale-ambiguous, inconsistent across views, and inaccurate near object boundaries. 
More importantly, independent pixels do not explicitly describe the extent, continuity, or complexity of the physical surface to which they belong. 
Consequently, conventional pixel-wise regularization~\cite{li2025geosvr,yan2024svraster} cannot reliably distinguish a coherent sparsely observed surface from isolated prediction errors.

We argue that coherent 2D surface regions provide a better interface between image-space priors and dense 3D geometry. 
Based on this insight, we introduce \textbf{SurfSVR}, a novel sparse voxel reconstruction paradigm that treats \textbf{2D surface priors as explicit 3D geometric regularizers}. 
SurfSVR first constructs surface regions by refining appearance-based superpixels with depth, normals, semantic boundaries, and cross-view geometric consistency. 
Each region is then represented by an adaptively selected planar or quadratic inverse-depth model according to its fitting reliability and geometric complexity, as shown in Figure~\ref{fig:teaser}. 
Regions that cannot be reliably explained by either low-order model remain available as complex or unknown regions rather than being forced into an unreliable fit.

The resulting structured priors are lifted into 3D and integrated throughout sparse voxel reconstruction based on GeoSVR~\cite{li2025geosvr}.
First, surface classes guide voxel subdivision and pruning: smooth planar regions stop at coarser octree levels, while complex surfaces and boundaries retain finer capacity. 
Second, the priors provide confidence-weighted depth, continuous subpixel depth, normal, and coverage supervision, enabling reliable surfaces to be optimized even when their initial opacity is low.
Third, a late-stage consensus filter removes only weak voxels that several reliable views place in free space, preserving supported geometry.

Unlike methods that use geometric predictions only as auxiliary pixel-wise losses, SurfSVR transforms heterogeneous 2D predictions into persistent surface-level constraints for voxel optimization and topology control.
As shown in Figure~\ref{fig:teaser}, SurfSVR produces more complete surfaces on weakly textured regions and better preserves fine geometric details than state-of-the-art methods.
Extensive experiments on three public benchmarks demonstrate consistent improvements across diverse geometry and visibility conditions, achieving state-of-the-art reconstruction quality.

Our contributions are summarized as follows:
\begin{itemize}
    \item We introduce a 2D-to-3D surface regularization paradigm that represents coherent image regions as explicit geometric constraints for sparse voxel reconstruction.
    \item We develop a geometry-aware surface construction method that combines appearance, depth, normals, semantic boundaries, and cross-view consistency with adaptive planar-or-quadratic inverse-depth fitting.
    \item We integrate the resulting priors into confidence-weighted dense and subpixel supervision, normal alignment, coverage control, and surface-aware topology refinement.
    \item SurfSVR achieves state-of-the-art reconstruction quality across three public benchmarks.
\end{itemize}

\section{Related Work}
\label{sec:related}

\subsection{Radiance Field Representations}
NeRF establishes differentiable volume rendering for learning radiance fields from posed images~\cite{mildenhall2021nerf}. Explicit and hybrid successors accelerate it with hash grids, tensor factorization, anti-aliased grids, and directly optimized sparse voxels~\cite{muller2022instant,chen2022tensorf,hu2023tri,liu2020nsvf,yu2021plenoctrees,yu2021plenoxels,sun2022direct}. In parallel, 3DGS enables efficient splatting with explicit Gaussian primitives, including structured and level-of-detail variants~\cite{kerbl20233dgs,lu2024scaffold,ren2024octree,yu2024mip}. SVRaster instead combines adaptive sparse voxels with rasterization~\cite{yan2024svraster}, and GeoSVR adapts it to surface reconstruction using depth constraints and voxel regularization~\cite{li2025geosvr}. These representations improve efficiency and scalability, but their primitives do not explicitly encode coherent physical surfaces. Consequently, voxel allocation and topology still rely mainly on local photometric and visibility evidence. SurfSVR uses coherent surface regions to regulate both geometry optimization and topology.

\subsection{Surface Reconstruction with Geometric Priors}
Classical methods reconstruct surfaces from multi-view correspondence and visibility~\cite{hartley2003multiple,furukawa2009accurate,shen2013accurate,zheng2014patchmatch,schoenberger2016sfm,schoenberger2016mvs,yao2018mvsnet}. Neural approaches learn continuous SDFs through differentiable rendering, with grids or staged optimization improving detail and efficiency~\cite{park2019deepsdf,yariv2020idr,oechsle2021unisurf,yariv2021volume,wang2021neus,wu2023voxurf,li2023neuralangelo,wang2024neurodin}. Gaussian methods instead align primitives with surfaces, adopt surfel parameterizations, derive opacity fields, or couple Gaussians with SDFs~\cite{guedon2024sugar,huang20242dgs,dai2024gsurfel,yu2024gof,yu2024gsdf,lyu20243dgsr,chen2023neusg,xu2024gsurf,zhang2024gspull}. Planar constraints, novel-view stereo, and local structural hints further improve mesh recovery~\cite{chen2024pgsr,wolf2024gs2mesh,wu2024surfacelocalhints}. Despite these advances, geometry remains optimized at the sample or primitive level without persistent region support.

Geometric priors mitigate photometric ambiguity through patch warping, multi-view features, or predicted depth and normals~\cite{darmon2022NeuralWarp,chen2023recovering,ren2024improving,yu2022monosdf,wang2022neuris,fu2022geoneus}. Gaussian and voxel methods similarly use normalized depth, depth-normal consistency, or uncertainty-aware monocular constraints~\cite{li2024dngaussian,dai2024vcrgaus,turkulainen2025dnsplatter,yu2024monogsdf,li2025geosvr}. Such per-pixel supervision remains vulnerable to scale ambiguity, boundary noise, and cross-view inconsistency. SurfSVR instead forms confidence-aware surface regions and lifts their extent, complexity, and support into persistent 3D constraints.

\begin{figure}[t]
\begin{center}
\includegraphics[width=\linewidth]{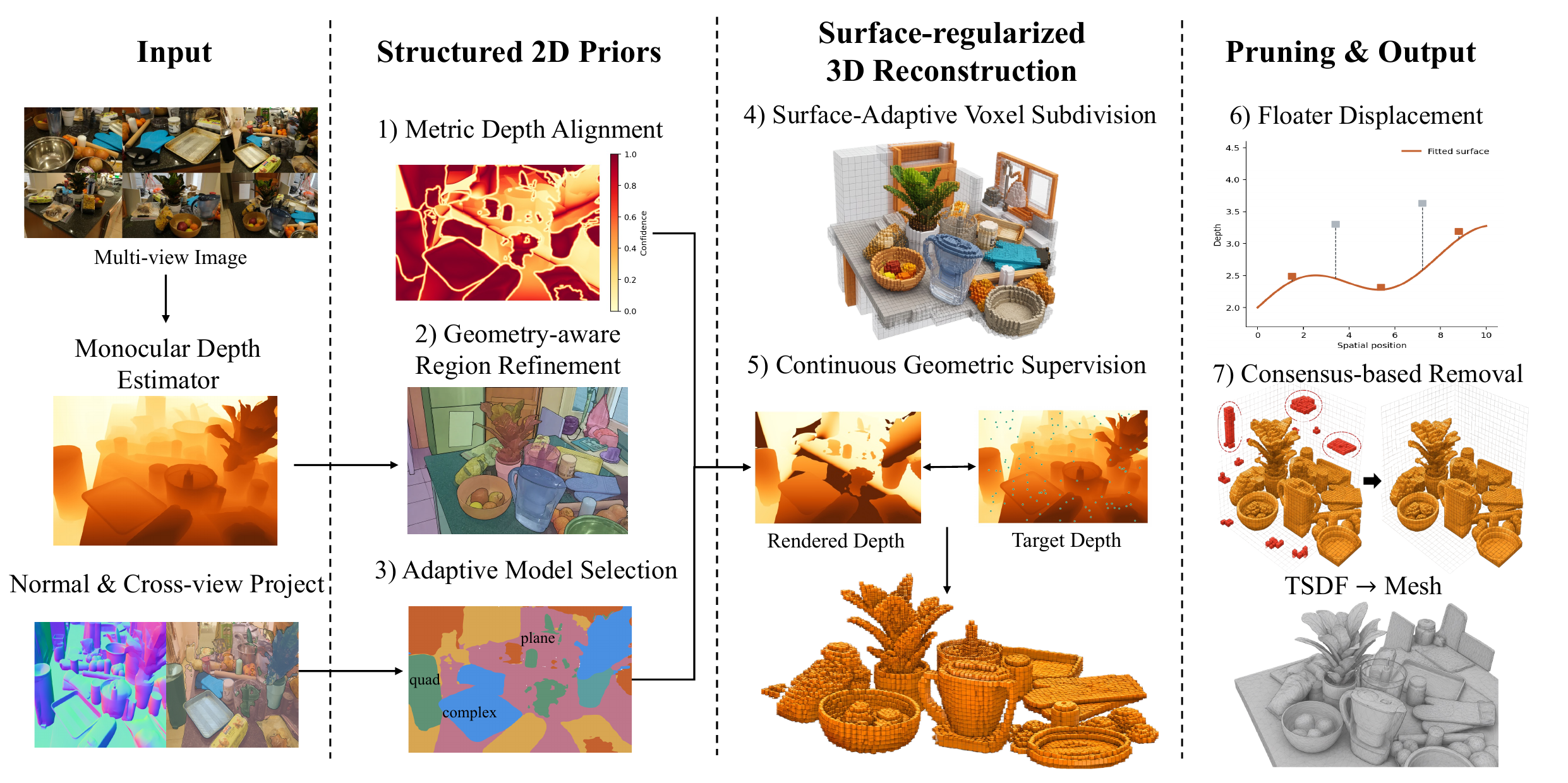}
\end{center}
\caption{
\textbf{Overview of the SurfSVR pipeline.} Given calibrated multi-view images and geometric priors, we refine appearance-based superpixels with depth, normals, semantic boundaries, and cross-view evidence, then fit each region with the simplest reliable planar or quadratic inverse-depth model. The fitted priors provide continuous geometric supervision and guide voxel subdivision, pruning, and conservative floater removal.
}
\label{fig:pipeline}
\end{figure}

\section{Method}

\subsection{Preliminaries:} Our SurfSVR framework is built upon SVRaster~\cite{yan2024svraster} and GeoSVR~\cite{li2025geosvr}. 
We first briefly review the basic steps in these two sparse voxel reconstruction methods.

\textbf{Scene Representation.}
Scenes are represented by Octree.
A voxel at Octree level $l \in [1, L]$ with grid index $v = \{i, j, k\}$ has size $v_s = w_s \cdot 2^{-l}$ and center $v_c = w_c - 0.5 \cdot w_s + v_s \cdot v$.
Each voxel stores Spherical Harmonics (SH) coefficients $v_{\text{sh}}$ for view-dependent color $c_i$, and $8$ corner density values $v_{\text{geo}} \in [0, +\infty)^{2 \times 2 \times 2}$ defining a continuous density field via trilinear interpolation $\text{interp}(\cdot)$.
Octree Voxels are dynamically pruned (low opacity) or subdivided (high gradient) during training.
 
\textbf{Volume Rendering.}
For a ray intersecting $N$ ordered voxels, pixel color $C$ is rendered via front-to-back $\alpha$-blending:
\begin{equation}
C = \sum_{i=1}^N T_i \alpha_i c_i, \quad T_i = \prod_{j=1}^{i-1} (1 - \alpha_j),
\end{equation}
where voxel opacity $\alpha_i = 1 - \exp\left(-\frac{\Delta t}{K} \sum_{k=1}^K \text{interp}(v_{\text{geo}}, q_k)\right)$ for $K$ samples along segment $\Delta t$ inside the voxel.
Depth $D$ is rendered similarly, while voxel surface normal $n$ is derived from the density gradient at center $q_c$:
\begin{equation}
n = \text{normalize}\left(\nabla_q \text{interp}(v_{\text{geo}}, q_c)\right).
\end{equation}

\subsection{Overview of SurfSVR}

Given COLMAP-calibrated multi-view images $\{\mathbf{I}_i\}_{i=1}^{N}$ and their depth/normal priors, SurfSVR reconstructs a sparse voxel field $\mathcal{V}$ whose voxels encode geometry, opacity, and view-dependent appearance.
Our core idea is to construct coherent surface regions in 2D and lift them into persistent 3D regularizers. Unlike pixel-wise supervision, a SurfSVR prior describes the spatial extent, geometric complexity, reliability, and cross-view support of a surface.

As shown in Figure~\ref{fig:pipeline}, we first construct 2D priors by refining appearance-based superpixels using depth, normal and cross-view geometric cues, then fit each region with an automatically selected surface model. 
The fitted priors are used directly in the optimization losses, while a voxel-adaptation module projects them into 3D to control subdivision, pruning, and conservative floater suppression.

\subsection{Structured 2D Surface Priors}
\paragraph{Geometry-aware region refinement.}

We initialize each image with appearance-based superpixels~\cite{gauen2025bist}, then construct a four-neighbor pixel graph whose edges are interrupted by relative-depth jumps, normal discontinuities, semantic boundaries, and depth-validity transitions. 
Connected components form the initial surface candidates.

Large candidates that cannot be explained by a low-order surface are recursively divided using spatial position, inverse depth, and normal features.
Adjacent fragments are merged if their shared boundary is geometrically weak and the union remains representable by a reliable surface model.

To enhance cross-view consistency, we unproject depth samples from nearby calibrated views and project them into the target image with per-source z-buffers.
Samples inconsistent with the target depth are rejected, and the remaining samples are confidence-weighted and fused into a cross-view depth field.
This field provides sub-pixel evidence for the current image and is used to refine region topology without copying every projected sample as a hard boundary.

\paragraph{Adaptive surface model selection.}
Given a refined surface region $\mathcal{R}$, SurfSVR models the inverse depth $s_{\mathcal{R}}(\mathbf{p})=z^{-1}$
of every pixel $\mathbf{p}=[u,v]^{\top}$ as a low-order polynomial defined on normalized image
coordinates:
\begin{equation}
s_{\mathcal{R}}(\mathbf{p})
=
\boldsymbol{\theta}_{\mathcal{R}}^{\top}
\boldsymbol{\phi}_{k}(\tilde{\mathbf{p}}),
\qquad
\tilde{\mathbf{p}}
=
\frac{\mathbf{p}-\boldsymbol{\mu}_{\mathcal{R}}}
{\boldsymbol{\sigma}_{\mathcal{R}}},
\end{equation}
where $\boldsymbol{\mu}_{\mathcal{R}}$ is the coordinate-wise median and
$\boldsymbol{\sigma}_{\mathcal{R}}$ is a robust coordinate scale computed
from the $90$th percentile absolute deviation within region $\mathcal{R}$,
clamped to at least one pixel, and
$\tilde{\mathbf{p}}=[\tilde{u},\tilde{v}]^{\top}$ denotes the normalized coordinate.
The vector
$\boldsymbol{\theta}_{\mathcal{R}}$
contains the polynomial coefficients to be estimated for region
$\mathcal{R}$, while
$\boldsymbol{\phi}_{k}(\cdot)$ is the basis function of a polynomial with
degree $k$.
We consider two models:
$
\boldsymbol{\phi}_{1}(x,y)
=
[1,x,y]^{\top},
$
which represents a planar surface, and
$
\boldsymbol{\phi}_{2}(x,y)
=
[1,x,y,x^{2},xy,y^{2}]^{\top},
$
which models smoothly curved surfaces.
Under the pinhole camera model, the first-order polynomial corresponds to a
3D plane in image space, whereas the second-order polynomial captures local
quadratic surface variations. 

SurfSVR first attempts the planar model and
introduces the quadratic model only when the former cannot satisfy the fitting
criteria, thereby selecting the simplest model that adequately explains the
geometry of the region.

We initialize each fit with RANSAC and refine using Tukey-biweight IRLS:
\begin{equation}
    \boldsymbol{\theta}_{\mathcal{R}}^{*}
    =
    \arg\min_{\boldsymbol{\theta}}
    \sum_{j\in\mathcal{R}}
    w_j
    \rho\left(
    \boldsymbol{\theta}^{\top}\boldsymbol{\phi}_{k}(\tilde{\mathbf{p}}_j)
    - z_j^{-1}
    \right).
\end{equation}
SurfSVR selects the simplest model satisfying fitting-error, inlier-ratio, support, positivity, and conditioning requirements. A planar model is tested first; the quadratic model is introduced only when necessary. Regions that cannot be reliably represented by either are retained as complex or unknown rather than being assigned a misleading low-order surface.

Region confidence $c_{\mathcal{R}}$ combines inlier ratio $r_{\mathcal{R}}$, fitting error $e_{\mathcal{R}}$, and support $n_{\mathcal{R}}$:
\begin{equation}
    c_{\mathcal{R}}
    =
    r_{\mathcal{R}}
    \exp\left(-\frac{e_{\mathcal{R}}}{\tau_e}\right)
    \min\left(1,\frac{n_{\mathcal{R}}}{\tau_n}\right),
\end{equation}
where $\tau_n$ is the support reference (128 samples in our configuration).

\subsection{Surface-Adaptive Sparse Voxels}

We lift 2D priors into 3D by projecting each voxel center into all available views. A voxel receives a smooth-surface vote when it lies within a voxel-scaled distance of a reliable planar or quadratic fit. Projections onto unreliable regions or boundaries produce complex-surface votes. Cross-view voting assigns each voxel a persistent class:
\begin{equation}
    \ell(\mathbf{x})
    \in
    \{\text{plane},\text{quadratic},\text{complex},\text{unknown}\}.
\end{equation}

The surface class regulates maximum octree level, subdivision priority, and conditional pruning protection. Planar regions stop at coarser levels, quadratic regions receive intermediate resolution, and complex or unknown regions retain the full voxel budget. A voxel can subdivide only if $l(\mathbf{x}) < L_{\ell(\mathbf{x})}$, where $L_{\ell}$ is the class-dependent level cap. Complex voxels receive higher subdivision priority, while reliable multi-view observations lower the risk of premature pruning. During late refinement, the topology is frozen and the surface losses continue to optimize geometry.

\subsection{Surface-Regularized Optimization}

Let $\alpha_i(\mathbf{p})=1-T_i(\mathbf{p})$ be the accumulated opacity of a rendered ray and $D_i(\mathbf{p})$ its accumulated depth numerator. We compute expected depth as
\begin{equation}
    \bar{z}_i(\mathbf{p})
    =
    \frac{D_i(\mathbf{p})}
         {\operatorname{sg}\!\left(\max(\alpha_i(\mathbf{p}),\epsilon)\right)},
\end{equation}
where $\operatorname{sg}(\cdot)$ denotes stop-gradient. This prevents the model from reducing geometric loss merely by changing opacity.
The complete surface regularization is
\begin{equation}
\begin{split}
    \mathcal{L}_{\mathrm{surf}}
    ={}&
    \eta(t)
    \left(
    \lambda_d\mathcal{L}_{d}
    +\lambda_s\mathcal{L}_{\mathrm{sub}}
    +\lambda_n\mathcal{L}_{n}
    \right)\\
    &+
    \eta_c(t)\lambda_c\mathcal{L}_{\mathrm{cov}},
\end{split}
\label{complete}
\end{equation}
where $\eta(t)$ gradually adds geometric supervision and $\eta_c(t)$ decays coverage supervision near the end of optimization.
For reliable pixel centers with sufficient rendered opacity, we employ a confidence-weighted robust log-depth loss:
\begin{equation}
    \mathcal{L}_{d}
    =
    \frac{
    \sum_{\mathbf{p}}c_{\mathcal{R}(\mathbf{p})}
    \rho_{\epsilon}\!\left(
    \log\bar{z}(\mathbf{p})-\log\hat{z}(\mathbf{p})
    \right)}
    {\sum_{\mathbf{p}}c_{\mathcal{R}(\mathbf{p})}},
\end{equation}
where $\rho_{\epsilon}(x)=\sqrt{x^2+\epsilon^2}$ denotes a robust distance function. Pixels are selected only when the fitted region passes the reliability gates and the rendered opacity exceeds a small depth-supervision floor; the depth denominator is detached from the gradient so that opacity alone cannot reduce the loss.

The fitted priors provide supervision at arbitrary continuous coordinates. We sample a continuous image coordinate $\mathbf{q}$ inside a reliable region:
\begin{equation}
    \mathbf{q}
    =
    \mathbf{p}+\boldsymbol{\delta},
    \qquad
    \boldsymbol{\delta}\sim\mathcal{U}(-0.5,0.5)^2.
\end{equation}
We normalize $\mathbf{q}$ using the same region-specific transform used during surface fitting:
\begin{equation}
    \tilde{\mathbf{q}}
    =
    \frac{\mathbf{q}-\boldsymbol{\mu}_{\mathcal{R}}}
    {\boldsymbol{\sigma}_{\mathcal{R}}}.
\end{equation}
The target depth at $\mathbf{q}$ is evaluated as
\begin{equation}
    z_{\mathcal{R}}(\mathbf{q})
    =
    \left[
    \boldsymbol{\theta}_{\mathcal{R}}^{\top}
    \boldsymbol{\phi}_{k_{\mathcal{R}}}
    \left(\tilde{\mathbf{q}}\right)
    \right]^{-1}.
\end{equation}
We bilinearly sample the rendered depth at $\mathbf{q}$ to compute
$\mathcal{L}_{\mathrm{sub}}$. Regions are sampled with inverse-area weighting
so that small surfaces are not overwhelmed.
The normal loss aligns rendered camera-space normals with the region prior:
\begin{equation}
    \mathcal{L}_{n}
    =
    \frac{\sum_{\mathbf{p}}c_{\mathcal{R}(\mathbf{p})}
    \left(1-
    \left|
    \mathbf{n}(\mathbf{p})^{\top}
    \hat{\mathbf{n}}(\mathbf{p})
    \right|\right)}
    {\sum_{\mathbf{p}}c_{\mathcal{R}(\mathbf{p})}}.
\end{equation}

Depth supervision alone provides little gradient for rays with very low
accumulated opacity, making it difficult to recover valid yet transparent
surfaces. To encourage reliable surface regions to acquire sufficient opacity,
we introduce a surface coverage loss:
\begin{equation}
\mathcal{L}_{\mathrm{cov}}
=
\frac{
\sum_{\mathbf{p}}
\tilde{c}(\mathbf{p})
\left[
\max\!\left(0,\alpha^{*}-\alpha(\mathbf{p})\right)
\right]^2}
{\sum_{\mathbf{p}}\tilde{c}(\mathbf{p})},
\end{equation}
where $\alpha^{*}$ is the target
minimum opacity threshold, and $\tilde{c}(\mathbf{p})$ is the confidence weight
used for coverage supervision. 
Specifically,
$\tilde{c}(\mathbf{p})=c_{\mathcal{R}}(\mathbf{p})$ only when pixel
$\mathbf{p}$ belongs to a reliable planar or quadratic surface region, or when
its geometry is independently verified by projected observations from
neighboring views; otherwise, $\tilde{c}(\mathbf{p})=0$.

During optimization, the coverage sampler explicitly includes the lowest-opacity
reliable pixels so that valid but under-reconstructed surfaces receive stronger
opacity gradients. Since excessive opacity may lead to artificially thick
surfaces, the weight of $\mathcal{L}_{\mathrm{cov}}$ is gradually reduced during
the late training stage, allowing the optimization to focus on accurate surface
positions rather than surface thickness.

\subsection{Surface-Aware Pruning and Floater Suppression}

Visibility alone cannot determine whether a weak voxel belongs to a valid sparsely observed surface. SurfSVR therefore lowers the pruning threshold for voxels supported by reliable planar or quadratic priors. This protection is deliberately conditional: a child with no rendering evidence can still be removed.

After the standard subdivision and pruning operations, we perform a separate
conservative floater-suppression step. For a voxel $\mathbf{x}$, let
$\mathcal{V}(\mathbf{x})$ denote the set of reliable views in which
$\mathbf{x}$ projects inside a valid fitted surface region and is visible.
For each $i\in\mathcal{V}(\mathbf{x})$, we define
\begin{equation}
    \Delta_i(\mathbf{x})
    =
    z_i(\mathbf{x})
    -
    z_{\mathcal{R}_i}\!\left(\pi_i(\mathbf{x})\right),
\end{equation}
which measures the camera-space displacement between the voxel center and the
fitted surface in view $i$. We then accumulate surface-support and free-space
votes as
\begin{align}
    S(\mathbf{x})
    &=
    \sum_{i\in\mathcal{V}(\mathbf{x})}
    \mathbb{I}\!\left[
    |\Delta_i(\mathbf{x})|
    \leq
    \tau_s h_{\mathbf{x}}
    \right],\\
    F(\mathbf{x})
    &=
    \sum_{i\in\mathcal{V}(\mathbf{x})}
    \mathbb{I}\!\left[
    \Delta_i(\mathbf{x})
    <
    -\tau_f h_{\mathbf{x}}
    \right].
\end{align}

Under the camera-space depth convention, $\Delta_i(\mathbf{x})<0$ indicates
that the voxel lies between the camera and the fitted surface and is therefore
supported as free-space evidence. We remove a voxel only when it receives no
surface-support votes, receives sufficient free-space votes, and has weak
rendering evidence:
\begin{equation}
    \operatorname{remove}(\mathbf{x})
    =
    \mathbb{I}\!\left[
    S(\mathbf{x})=0
    \;\land\;
    F(\mathbf{x})\geq n_f
    \;\land\;
    e_{\mathrm{render}}(\mathbf{x})<\tau_r
    \right].
\end{equation}
Here, $e_{\mathrm{render}}(\mathbf{x})$ denotes the rendering-evidence score
used by the underlying pruning procedure, and lower values indicate weaker
support. We cap the number of removed voxels per pass. The topology is then
fixed during late refinement while continuous surface supervision remains active.

\begin{table}[t]
\caption{Quantitative comparison on DTU in terms of Chamfer distance ($\downarrow$).
Green and light-yellow cells denote the best and second-best results, respectively.}
\label{tab:dtu_results}
\begin{center}
\renewcommand{\arraystretch}{1.12}
\setlength{\tabcolsep}{3.2pt}
\resizebox{\textwidth}{!}{
\begin{tabular}{@{}lcccccccccccccccc}
\toprule
Method &
24 & 37 & 40 & 55 & 63 & 65 & 69 & 83 &
97 & 105 & 106 & 110 & 114 & 118 & 122 & Mean \\
\midrule
VolSDF~\cite{yariv2021volume}
& 1.14 & 1.26 & 0.81 & 0.49 & 1.25 & 0.70 & 0.72 & 1.29
& 1.18 & 0.70 & 0.66 & 1.08 & 0.42 & 0.61 & 0.55 & 0.86 \\

NeuS~\cite{wang2021neus}
& 1.00 & 1.37 & 0.93 & 0.43 & 1.10 & 0.65 & 0.57 & 1.48
& 1.09 & 0.83 & 0.52 & 1.20 & 0.35 & 0.49 & 0.54 & 0.84 \\

Neuralangelo~\cite{li2023neuralangelo}
& 0.37 & 0.72 & 0.35 & 0.35 & 0.87 & 0.54 & 0.53 & 1.29
& 0.97 & 0.73 & 0.47 & 0.74 & 0.32 & 0.41 & 0.43 & 0.61 \\

Geo-NeuS~\cite{fu2022geoneus}
& 0.38 & 0.54 & 0.34 & 0.36 & 0.80
& \best{0.45} & \best{0.41} & {1.03}
& 0.84 & {0.55} & 0.46 & 0.47
& \best{0.29} & 0.36 & 0.35 & 0.51 \\

MonoSDF~\cite{yu2022monosdf}
& 0.66 & 0.88 & 0.43 & 0.40 & 0.87 & 0.78 & 0.81 & 1.23
& 1.18 & 0.66 & 0.66 & 0.96 & 0.41 & 0.57 & 0.51 & 0.73 \\

2DGS~\cite{huang20242dgs}
& 0.48 & 0.91 & 0.39 & 0.39 & 1.01 & 0.83 & 0.81 & 1.36
& 1.27 & 0.76 & 0.70 & 1.40 & 0.40 & 0.76 & 0.52 & 0.80 \\

GOF~\cite{yu2024gof}
& 0.50 & 0.82 & 0.37 & 0.37 & 1.12 & 0.74 & 0.73 & 1.18
& 1.29 & 0.68 & 0.77 & 0.90 & 0.42 & 0.66 & 0.49 & 0.74 \\

SVRaster~\cite{yan2024svraster}
& 0.61 & 0.74 & 0.41 & 0.36 & 0.93 & 0.75 & 0.94 & 1.33
& 1.40 & 0.61 & 0.63 & 1.19 & 0.43 & 0.57 & 0.44 & 0.76 \\

GS2Mesh~\cite{wolf2024gs2mesh}
& 0.59 & 0.79 & 0.70 & 0.38 & 0.78 & 1.00 & 0.69 & 1.25
& 0.96 & 0.59 & 0.50 & 0.68 & 0.37 & 0.50 & 0.46 & 0.68 \\

VCR-GauS~\cite{dai2024vcrgaus}
& 0.55 & 0.91 & 0.40 & 0.43 & 0.97 & 0.95 & 0.84 & 1.39
& 1.30 & 0.90 & 0.76 & 0.92 & 0.44 & 0.75 & 0.54 & 0.80 \\

MonoGSDF~\cite{yu2024monogsdf}
& 0.45 & 0.65 & 0.36 & 0.36 & 0.94 & 0.70 & 0.67 & 1.27
& 0.99 & 0.63 & 0.49 & 0.84 & 0.39 & 0.53 & 0.47 & 0.65 \\

PGSR~\cite{chen2024pgsr}
& 0.36 & 0.57 & 0.38 & \second{0.33} & 0.78 & 0.58 & 0.50 & 1.08
& 0.63 & 0.59 & 0.46 & 0.54 & 0.30 & 0.38 & 0.34 & 0.52 \\

AmbiSuR~\cite{revisiting}
& \second{0.32} & \best{0.48} & {0.31}
& \second{0.33} & \best{0.65} & 0.48 & \second{0.42}
& \best{0.97} & \best{0.61} & \best{0.54}
& {0.36} & \best{0.44} & \best{0.29}
& \second{0.34} & {0.34} & \second{0.46} \\

GeoSVR~\cite{li2025geosvr}
& \second{0.32} & {0.51} & \second{0.30}
& \second{0.33} & {0.71} & 0.48 & \second{0.42}
& {1.03} & \second{0.62} & 0.56
& \best{0.33} & \second{0.46} & {0.30}
& \second{0.34} & \second{0.32} & {0.47} \\

\midrule
\textbf{SurfSVR (Ours)}
& \best{0.31} & \second{0.49} & \best{0.29} & \best{0.30}
& \best{0.65} & \best{0.45} & {0.43} & \best{0.97}
& \second{0.62} & \best{0.54} & \second{0.35} & {0.47}
& {0.30} & \best{0.33} & \best{0.31} & \best{0.45} \\
\bottomrule
\end{tabular}
}
\end{center}
\end{table}

\begin{figure}[h]
\begin{center}
\includegraphics[width=0.8\linewidth]{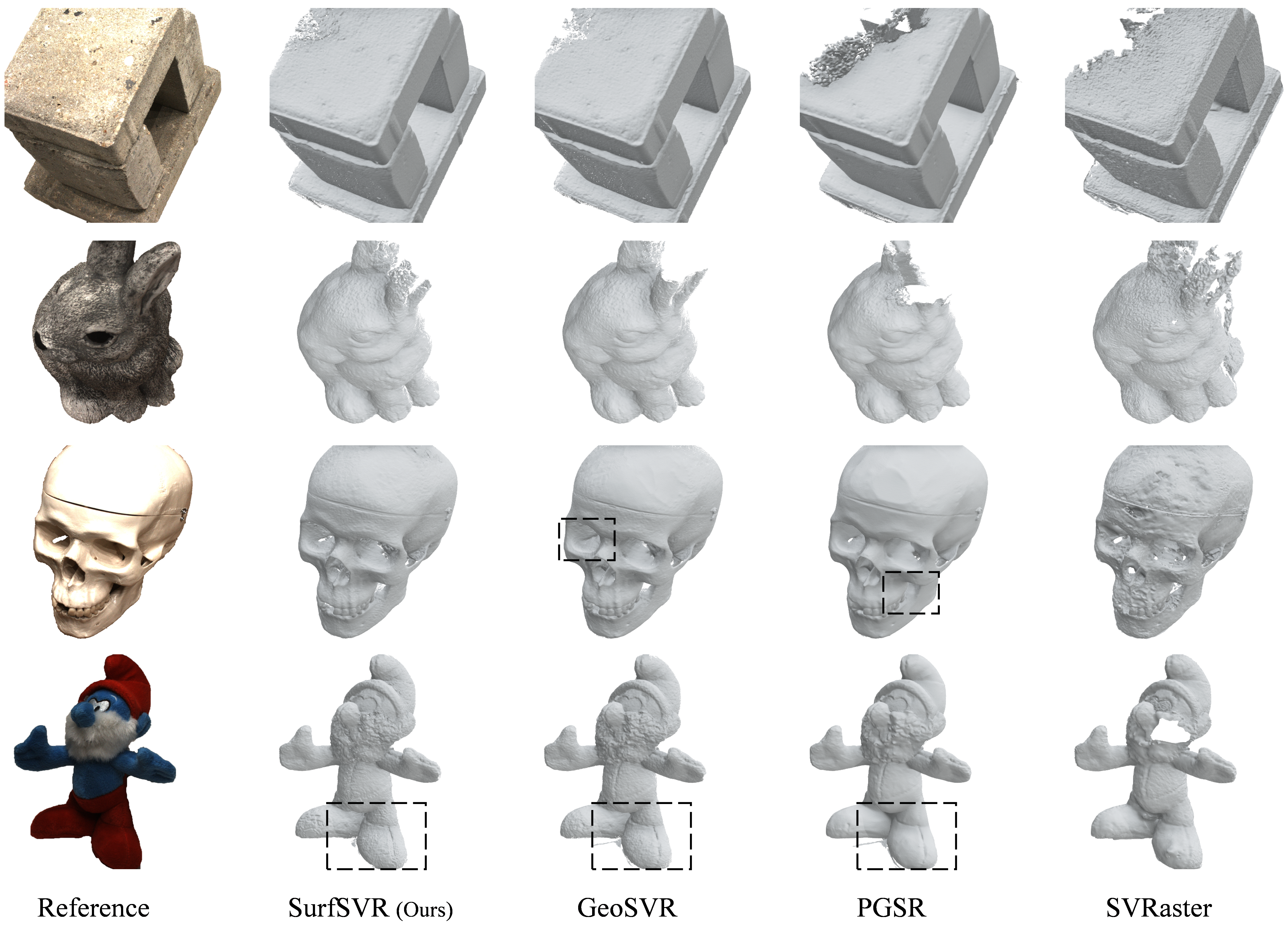}
\end{center}
\caption{
\textbf{Visualization results on DTU dataset.}}
\label{fig:resfig}
\end{figure}

\section{Experiments}

\subsection{Experimental Setup}

\paragraph{Datasets.}
Following GeoSVR~\cite{li2025geosvr}, we evaluate SurfSVR on three public benchmarks: DTU~\cite{jensen2014large}, Tanks and Temples (TnT)~\cite{knapitsch2017tanks}, and Mip-NeRF 360~\cite{barron2022mipnerf360}. 
For DTU, we use the 15 commonly evaluated scans and report Chamfer Distance (lower is better). 
For TnT, we evaluate on 5 scenes from the training split and report F1-Score (higher is better). 
Courthouse is removed due to inaccurate ground truth.
For Mip-NeRF 360, we use all nine scenes (five outdoor and four indoor scenes) and evaluate novel-view synthesis quality. 
We follow the same image downsampling, camera preprocessing, scene selection, and evaluation protocols as GeoSVR.

\paragraph{Implementation details.}
SurfSVR is implemented upon the GeoSVR. 
On DTU, we first optimize the complete SurfSVR using Equation~\ref{complete} and apperance loss from GeoSVR for 20,000 iterations, then conduct 2,000 iterations of topology-frozen surface-guided late-refinement.
On TnT and Mip-NeRF, the training step is 20,000 for complete training and 10,000 for late-refinement.
Structured priors are precomputed offline using BiST superpixels, neighboring 48-view depth projection, and robust surface fitting.
Training uses reliable regions selected by degree, confidence, fitting-error, inlier, support, and opacity gates.
For mesh extraction, we adopt TSDF fusion with a voxel size of $0.002$ on DTU and follow PGSR~\cite{chen2024pgsr} to determine extraction resolution for TnT.
Surface losses are gradually activated after initialization. 
Coverage supervision $\mathcal{L}_{\mathrm{cov}}$ is decayed linearly to 0 during late-refinement to prevent surface thickening.
Other details are provided in supplementary material.
All experiments are conducted on a single NVIDIA A100 GPU.

\subsection{Comparison with the State of the Art}

\paragraph{DTU.}
Table~\ref{tab:dtu_results} compares SurfSVR with baseline methods across all 15 DTU scans. 
SurfSVR achieves the best overall performance with a mean Chamfer distance of \textbf{0.45}, outperforming both GeoSVR (0.47) and AmbiSuR (0.46).
Notably, SurfSVR obtains the best results, including ties, on 9 out of 15 scenes and the marked second-best results on 3 scenes.
It matches the best baseline on scan 63 and achieves a tied best result on scan 65, two challenging scenes with weakly textured surfaces. 
Figure~\ref{fig:resfig} visualizes the reconstructed meshes, demonstrating that SurfSVR produces more complete surfaces on smooth regions while preserving fine geometric details better than baseline methods.

\begin{table}[t]
\caption{
Quantitative comparison on Tanks and Temples.
We report per-scene F1-score ($\uparrow$) and mean F1-score.
* denotes results of our own re-implementation under the same experimental conditions.
}
\label{tab:tnt_results}
\begin{center}
\renewcommand{\arraystretch}{1.14}
\small
\begin{tabular*}{\textwidth}{
@{\extracolsep{\fill}}
lcccccc}
\toprule
Method &
Barn &
Caterpillar &
Ignatius &
Meetingroom &
Truck &
Mean \\
\midrule
NeuS~\cite{wang2021neus}
& 0.29 & 0.29 & \second{0.83} & 0.24 & 0.45 & 0.42 \\

Neuralangelo~\cite{li2023neuralangelo}
& \best{0.70} & 0.36 & \best{0.89} & 0.32 & 0.48 & 0.55 \\

Geo-NeuS~\cite{fu2022geoneus}
& 0.33 & 0.26 & 0.72 & 0.20 & 0.45 & 0.39 \\

MonoSDF~\cite{yu2022monosdf}
& 0.49 & 0.31 & 0.78 & 0.23 & 0.42 & 0.45 \\

2DGS~\cite{huang20242dgs}
& 0.41 & 0.23 & 0.51 & 0.17 & 0.45 & 0.35 \\

GOF~\cite{yu2024gof}
& 0.51 & 0.41 & 0.68 & 0.28 & 0.59 & 0.49 \\

SVRaster~\cite{yan2024svraster}
& 0.35 & 0.33 & 0.69 & 0.19 & 0.54 & 0.42 \\

VCR-GauS~\cite{dai2024vcrgaus}
& 0.62 & 0.26 & 0.61 & 0.19 & 0.52 & 0.44 \\

MonoGSDF~\cite{yu2024monogsdf}
& 0.56 & 0.38 & 0.72 & 0.25 & 0.62 & 0.51 \\

PGSR~\cite{chen2024pgsr}
& 0.66 & 0.44 & 0.81 & 0.33
& {0.66} & 0.58 \\
\midrule
GeoSVR*~\cite{li2025geosvr}
& \second{0.68} & \second{0.49}
& \second{0.83} & {0.37}
& 0.65 & 0.60 \\

$\text{AmbiSuR}_\text{M}$~\cite{revisiting}
& 0.67 & \second{0.49}
& \second{0.83} & \second{0.39}
& \second{0.68} & \second{0.61} \\

\textbf{SurfSVR (Ours)}
& {0.67} & \best{0.52}
& \second{0.83} & \best{0.40} & \best{0.69}
& \best{0.62} \\
\bottomrule
\end{tabular*}
\end{center}
\end{table}

\begin{figure}[t]
\begin{center}
\includegraphics[width=\linewidth]{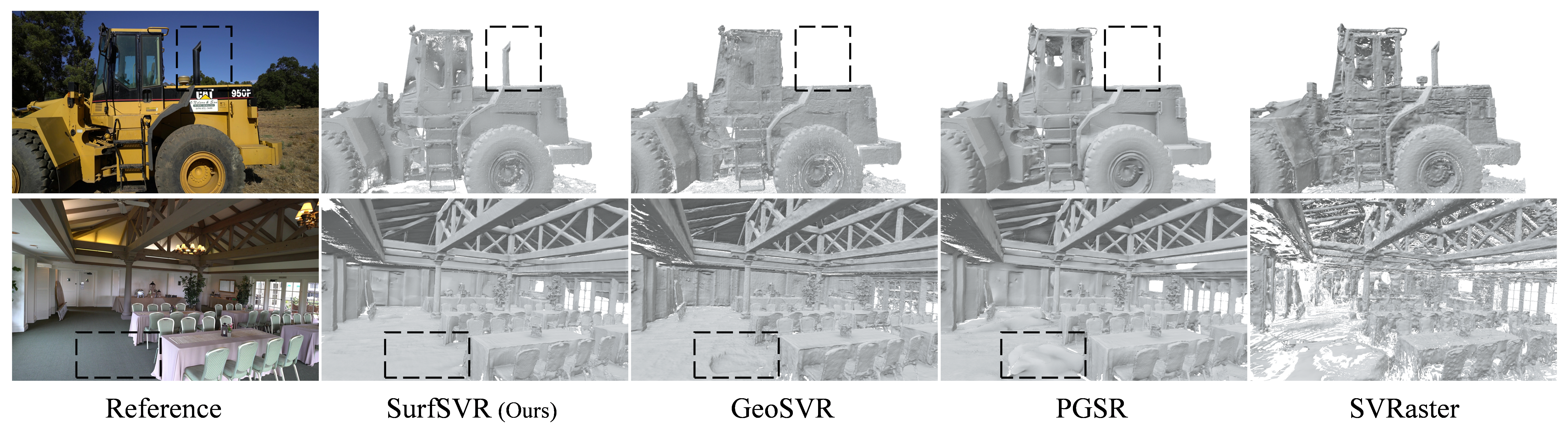}
\end{center}
\caption{
\textbf{Visualization results on TnT dataset.} Our results preserve more reliable geometry than baselines.}
\label{fig:resfig-TnT}
\end{figure}

\begin{table}[t]
\centering

\begin{minipage}[t]{0.67\linewidth}
\vspace{0pt}

\captionof{table}{
Quantitative comparison on Mip-NeRF 360.
Green and light-yellow cells denote the best and second-best results.
}
\label{tab:mip360_results}

\centering
\setlength{\tabcolsep}{2.5pt}
\renewcommand{\arraystretch}{1.05}

\resizebox{\linewidth}{!}{
\begin{tabular}{@{}lcccccc}
\toprule
& \multicolumn{3}{c}{Outdoor} &
\multicolumn{3}{c}{Indoor} \\
\cmidrule(lr){2-4}\cmidrule(lr){5-7}
Method &
PSNR$\uparrow$ & SSIM$\uparrow$ & LPIPS$\downarrow$ &
PSNR$\uparrow$ & SSIM$\uparrow$ & LPIPS$\downarrow$ \\
\midrule

NeRF
& 21.46 & 0.458 & 0.515 & 26.84 & 0.790 & 0.370 \\

Deep Blending
& 21.54 & 0.524 & 0.364 & 26.40 & 0.844 & 0.261 \\

Instant NGP
& 22.90 & 0.566 & 0.371 & 29.15 & 0.880 & 0.216 \\

Mip-NeRF 360
& 24.47 & 0.691 & 0.283
& \best{31.72} & 0.917 & 0.180 \\

3DGS
& 24.67 & 0.728 & 0.240
& \second{30.96} & 0.924 & 0.187 \\

SVRaster
& 24.68 & 0.738 & 0.206
& 30.65 & 0.927 & \second{0.161} \\

BakedSDF
& 22.47 & 0.585 & 0.349
& 27.06 & 0.836 & 0.258 \\

SuGaR
& 22.93 & 0.629 & 0.356
& 29.43 & 0.906 & 0.225 \\

2DGS
& 24.34 & 0.717 & 0.246
& 30.40 & 0.916 & 0.195 \\

GOF
& 24.82 & 0.750 & \best{0.202}
& 30.79 & 0.924 & 0.184 \\

VCR-GauS
& 24.31 & 0.707 & 0.280
& 30.53 & 0.921 & 0.184 \\

PGSR
& 24.76 & \best{0.752} & \second{0.203}
& 30.36 & \best{0.934} & \best{0.147} \\

GeoSVR
& \second{24.83} & 0.738 & 0.218
& 30.46 & 0.921 & 0.172 \\
\midrule
\textbf{SurfSVR (Ours)}
& \best{24.90} & \second{0.750} & \second{0.203}
& 30.76 & \second{0.929} & 0.166 \\

\bottomrule
\end{tabular}
}

\end{minipage}
\hfill
\begin{minipage}[t]{0.32\linewidth}
\vspace{0pt}

\captionof{table}{
Runtime comparison of different surface reconstruction methods on DTU dataset.
}
\label{tab:runtime_comparison}

\centering
\setlength{\tabcolsep}{4pt}
\renewcommand{\arraystretch}{1.05}

\begin{tabular}{lc}
\toprule
\textbf{Method} & \textbf{Time (h)} \\
\midrule

NeuS & $>12$h \\
Neuralangelo & $>128$h \\
Geo-NeuS & $>12$h \\
MonoSDF & 6h \\

2DGS & 0.2h \\
GOF & 1h \\
GS2Mesh & 0.3h \\
VCR-GauS & 1h \\
PGSR & 0.5h \\
GeoSVR & 0.8h \\
AmbiSuR & 0.6h \\

\midrule
SurfSVR (Ours) & 0.9h \\

\bottomrule
\end{tabular}

\end{minipage}

\end{table}

\paragraph{Tanks and Temples.}
Table~\ref{tab:tnt_results} reports per-scene F1-scores on TnT. 
SurfSVR achieves the best mean F1-score of 0.62, surpassing GeoSVR (0.60) and mono-AmbiSuR (0.61). 
We achieve best performance on Caterpillar, Meetingroom and Truck scenes.
Figure~\ref{fig:resfig-TnT} demonstrates the visulization comparison results.
SurfSVR reliably recovers thin structures in Caterpillar as highlighted in rectangle.
For Meetingroom, our surface fitting strategy helps recover the reflective ground.

{}

\paragraph{Mip-NeRF 360.}
Since Mip-NeRF 360 does not provide ground-truth geometry, we follow previous work and evaluate novel-view synthesis quality, reporting results separately for outdoor and indoor scenes. Table~\ref{tab:mip360_results} shows that SurfSVR achieves competitive rendering quality while maintaining accurate geometry. For outdoor scenes, SurfSVR obtains the best PSNR (24.90) and second-best SSIM (0.750), comparable to PGSR and GOF. For indoor scenes, our method achieves second-best SSIM (0.929) and competitive LPIPS (0.166). These results demonstrate that enforcing geometric accuracy through surface priors does not compromise appearance reconstruction quality, validating the effectiveness of our surface-regularized optimization framework.

\begin{table}[h]
\caption{
Ablation study of SurfSVR. Stage I ablates 2D-to-3D constraints;
Stage II replaces the 2D surface construction under the full pipeline.
}
\label{tab:ablation}
\begin{center}
\renewcommand{\arraystretch}{1.08}
\setlength{\tabcolsep}{2pt}
\scriptsize
\begin{tabular}{@{}c p{0.62\columnwidth} cc@{}}
\toprule
ID & Variant &
\makecell[c]{DTU Chamfer\\$\downarrow$} &
\makecell[c]{TnT F1\\$\uparrow$} \\
\midrule

\rowcolor{GroupGray}
\multicolumn{4}{@{}l}{\textit{Stage I: 2D-to-3D constraints}} \\

A & GeoSVR baseline*
& 0.477 & 0.606 \\

B & A + continuous \newline
    surface supervision
& 0.465 & 0.617 \\

C & B + surface-adaptive \newline
    subdivision and pruning
& 0.464 & 0.619 \\

D & {C + Floater \newline suppression
    }
& 0.454 & 0.622 \\

\addlinespace[1pt]
\rowcolor{GroupGray}
\multicolumn{4}{@{}l}{\textit{Stage II: 2D surface construction}} \\

E & Full SurfSVR w/o \newline
    appearance-region fitting
& 0.461 & 0.615 \\

F & Full SurfSVR w/o \newline
    boundary-refined surfaces
& 0.458 & 0.620 \\

G & Full SurfSVR w/ \newline
    only planar surface
& 0.456 & 0.620 \\

H & \textbf{Full SurfSVR w/o cross-view \newline
    surface evidence (Ours)}
& {0.457} & {0.621} \\

\bottomrule
\end{tabular}
\end{center}
\end{table}

\subsection{Ablation Study}

We perform an incremental ablation study to evaluate both the effectiveness of the proposed 2D-to-3D surface regularization pipeline and the quality of the constructed surface priors in Table~\ref{tab:ablation}.

\paragraph{Stage I: Effectiveness of 2D-to-3D surface regularization.}

Starting from the GeoSVR baseline (Row A), we progressively introduce each component of the proposed surface-guided optimization pipeline.

Adding continuous surface supervision (Row B) reduces the DTU Chamfer distance from 0.477 to 0.465 and improves the TnT F1-score from 0.606 to 0.617. This demonstrates that region-level depth, continuous subpixel supervision, normal alignment, and coverage constraints provide substantially denser and more reliable geometric guidance than pixel-wise supervision alone.

Row C further incorporates surface-adaptive subdivision and surface-aware pruning. Although the numerical improvement is relatively modest (0.465$\rightarrow$0.464 on DTU and 0.617$\rightarrow$0.619 on TnT), these operations improve voxel allocation efficiency by assigning finer resolution only to geometrically complex regions while protecting reliable but sparsely observed surfaces from premature pruning.

Finally, Row D enables the proposed post-stage floater suppression strategy, yielding the largest gain in the second stage of optimization. 
DTU Chamfer further decreases from 0.464 to 0.454, while the TnT F1-score increases from 0.619 to 0.622. 
Since voxels are removed only when multiple reliable surface observations consistently identify them as free space, the proposed consensus mechanism effectively suppresses floating artifacts without sacrificing valid thin structures or reconstruction completeness.

\paragraph{Stage II: Quality of structured surface construction.}

Under the complete reconstruction pipeline, we further analyze how different surface construction strategies affect the final reconstruction quality.

Removing appearance-region fitting (Row E) leads to a noticeable performance degradation (0.454$\rightarrow$0.461 on DTU and 0.622$\rightarrow$0.615 on TnT), indicating that initializing surface candidates from coherent appearance regions provides an important structural prior before geometric refinement.

Row F restores appearance-region fitting but removes the proposed boundary refinement, resulting in partial recovery (0.458 on DTU and 0.620 on TnT). This shows that refining region topology using depth discontinuities, normal changes, and geometric boundaries substantially improves the quality of fitted surfaces by preventing different physical surfaces from being merged together.

Row G indicates that using planar 2D surface priors alone results in no performance degradation on the large-scale TnT and only a little drop on object-centric dataset DTU, since 67\% of the surface priors on DTU and 89\% on TnT are classified as planar surfaces in SurfSVR.

Finally, Row H further evaluates the effectiveness of cross-view geometry in surface prior construction. 
Without this term, the performance drops from 0.454 to 0.457 on DTU and from 0.622 to 0.621 on TnT, which is a small but consistent degradation in the reported ablation.
The experiment verifies its importance.

\subsection{Efficiency and Limitations}

SurfSVR requires approximately 54 minutes to process a DTU scene on a single NVIDIA A100 GPU, including superpixel extraction and robust cross-view surface fitting. 
Table~\ref{tab:runtime_comparison} compares SurfSVR against baselines.
The additional cost of SurfSVR comes from offline surface-prior construction and the added geometric supervision during voxel optimization.
SurfSVR trades part of the efficiency advantage of GeoSVR for more accurate geometry and more reliable surface topology. 
Future work may accelerate the pipeline through parallel surface fitting, incremental cross-view projection, and tighter coupling between the optimization of 2D priors and the 3D voxel representation.

\section{Conclusion}

We presented SurfSVR, a surface-regularized sparse voxel reconstruction framework that transforms structured 2D surface priors into persistent 3D geometric constraints. SurfSVR constructs reliable regions from appearance, depth, normals and cross-view consistency, and represents them with adaptively selected planar-quadratic models.
These priors directly supervise continuous geometry and regulate voxel subdivision, pruning, and floater suppression, allowing surface structure to influence both geometry optimization and topology. 
Experiments demonstrate state-of-the-art reconstruction accuracy on  3 public benchmarks.
Future work will focus on more efficient surface construction and tighter joint optimization.

\bibliography{iclr2027_conference}
\bibliographystyle{iclr2027_conference}

\end{document}